%% file: main.tex
\documentclass[10pt,twocolumn,letterpaper]{article}

\usepackage{cvpr}              

\input{preamble}

\definecolor{cvprblue}{rgb}{0.21,0.49,0.74}
\usepackage[pagebackref,breaklinks,colorlinks,citecolor=cvprblue]{hyperref}

\def\paperID{14412} 
\def\confName{CVPR}
\def\confYear{2024}

\title{AnimateAnything: Fine-Grained Open Domain Image Animation with Motion Guidance}

\author{Zuozhuo Dai 
,
Zhenghao Zhang
,
Yao Yao
,
Bingxue Qiu
,
Siyu Zhu
,
Long Qin
,
Weizhi Wang\\
Alibaba Group\\
}

\begin{document}
\maketitle
\input{sec/0_abstract}    
\input{sec/1_intro}
\input{sec/2_related}

\input{sec/3_method}

\input{sec/4_exp}
\input{sec/5_ablation}
\input{sec/6_conclusion}
{
    \small
    \bibliographystyle{ieeenat_fullname}
    \bibliography{main}
}


\end{document}

%% file: preamble.tex
\usepackage[dvipsnames]{xcolor}


%% file: sec/0_abstract.tex
\begin{abstract}

Image animation is a key task in computer vision which aims to generate dynamic visual content from a static image. Recent image animation methods employ neural based rendering technique to generate realistic animations. Despite these advancements, achieving fine-grained and controllable image animation guided by text remains challenging, particularly for open-domain images captured in diverse real environments. In this paper, we introduce an open domain image animation method that leverages the motion prior of video diffusion model. Our approach introduces targeted motion area guidance and motion strength guidance, enabling precise control the movable area and its motion speed. This results in enhanced alignment between the animated visual elements and the prompting text, thereby facilitating a fine-grained and interactive animation generation process for intricate motion sequences. We validate the effectiveness of our method through rigorous experiments on an open-domain dataset, with the results showcasing its superior performance. 
Project page can be found at \href{https://animationai.github.io/AnimateAnything/}{https://animationai.github.io/AnimateAnything}.

\end{abstract}

%% file: sec/1_intro.tex
\section{Introduction}\label{sec:intro}
\begin{figure*}
    \centering
    \includegraphics[width=\textwidth]{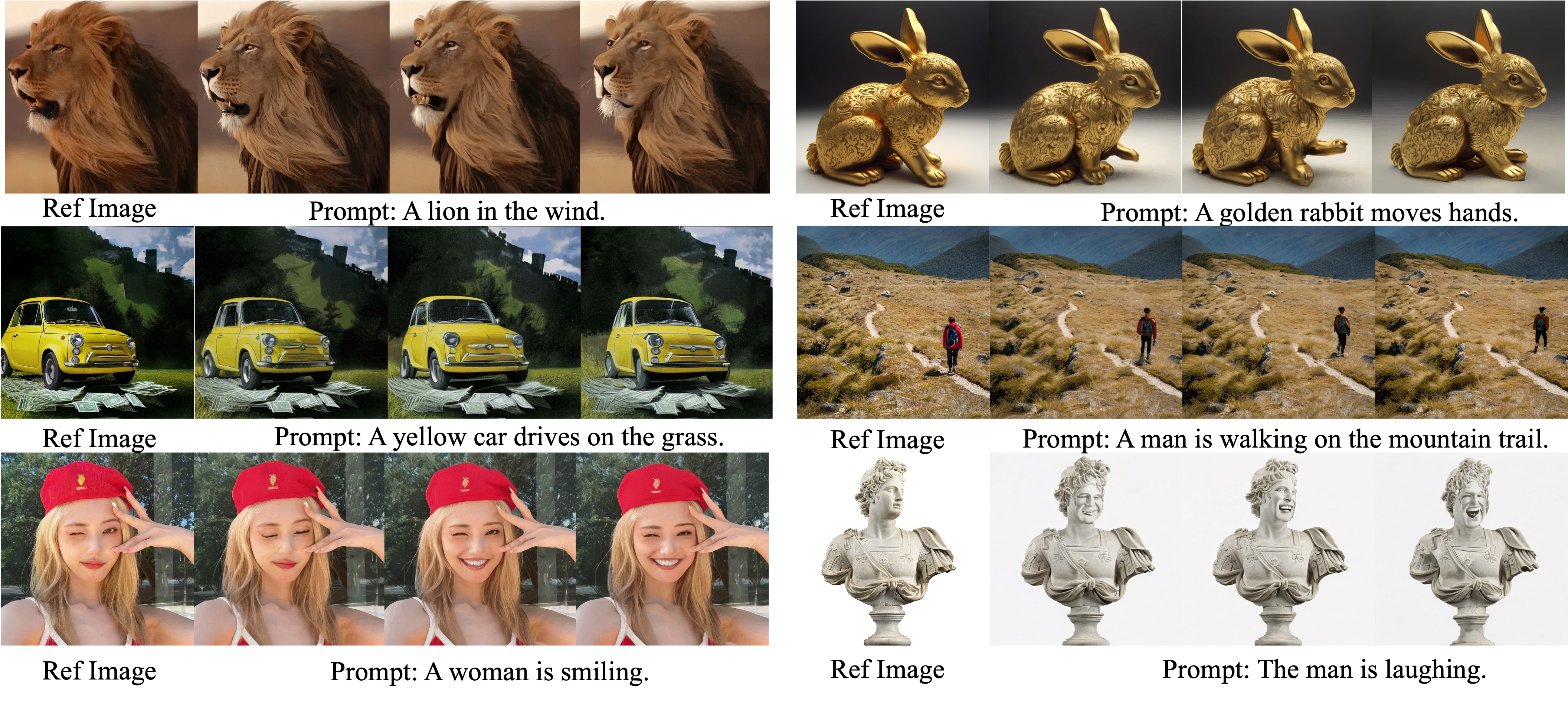}
    \includegraphics[width=\textwidth]{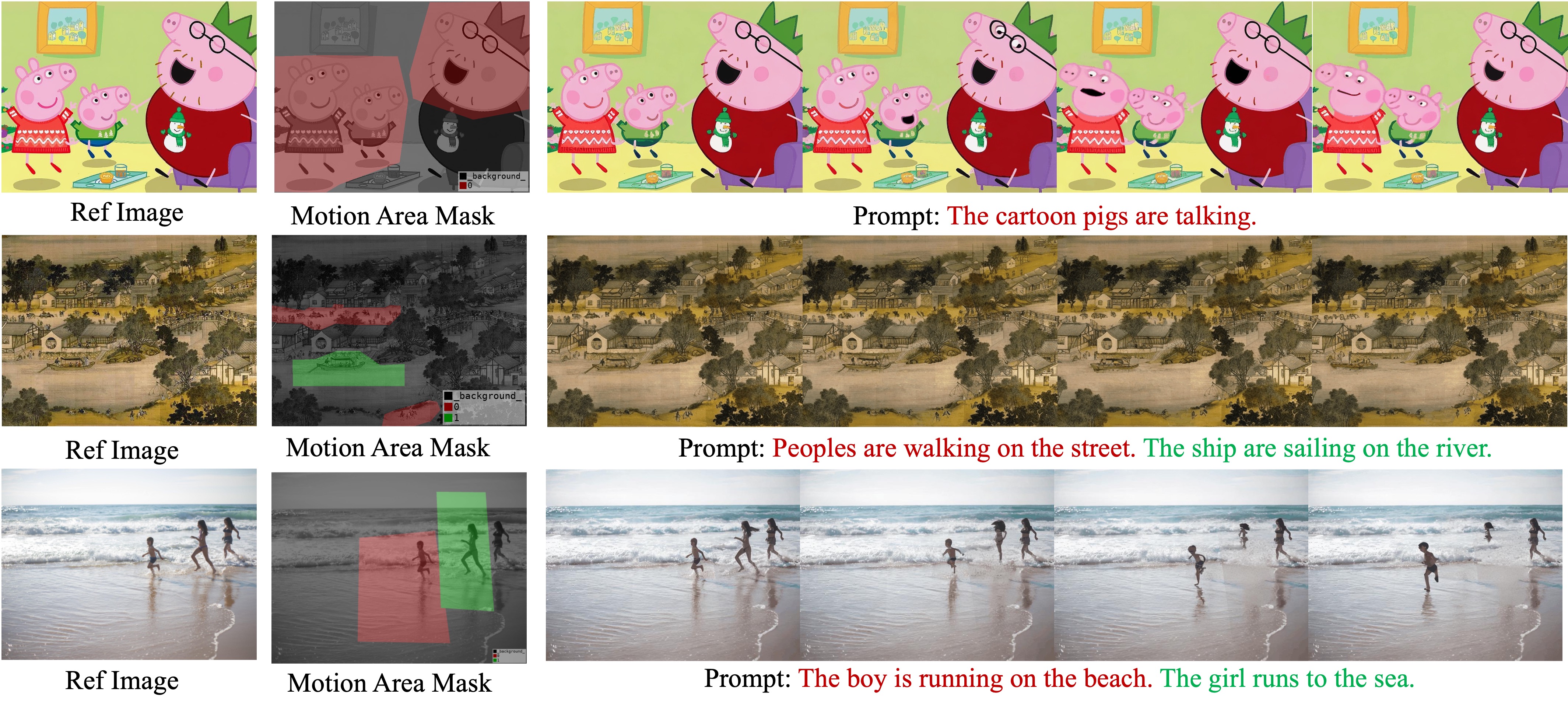}
    \vspace{-5mm}
    \caption{
    Examples of our method for image animation. The first three rows illustrate the use of prompt text to animate the reference image, with the 6th, 11th, and 16th frames of the generated animation visualized. The last three rows demonstrate the precise control of movable objects using motion mask guidance in images containing multiple objects. The last two rows show the iterative generation of animation using red and green prompts to animate objects in the corresponding masks. Additional examples are provided in the supplemental material.
    }
    \label{fig:it2v}   
    \vspace{-5mm}
\end{figure*}

The field of image animation has recently gained attention, especially for creating short videos and animated GIFs for social media and photo-sharing platforms. However, current image animation methods~\cite{endo2019animating,holynski2021animating,mahapatra2022controllable,Li20233DCF,bertiche2023blowing,Mahapatra2023SynthesizingAC} are limited in animating specific object types, such as fluid~\cite{holynski2021animating,mahapatra2022controllable, okabe2009animating,Li20233DCF,Mahapatra2023SynthesizingAC}, natural scenes~\cite{xiong2018learning,li2023generative,cheng2020time,jhou2015animating,shaham2019singan}, human hair~\cite{xiao2023automatic}, portraits~\cite{wang2021latent,wang2020imaginator,geng2018warp}, and bodies~\cite{wang2021latent,weng2019photo,karras2023dreampose,siarohin2021motion,blattmann2021understanding,bertiche2023blowing}, limiting their practical application in open domain scenarios. Recent advancements in video diffusion models~\cite{ho2022video, esser2023structure,wang2023videocomposer,singer2022make,khachatryan2023text2video, he2022latent,zhou2022magicvideo,luo2023videofusion} have enabled the generation of diverse and realistic videos based on reference texts and images. In this paper, we aim to address the open domain image animation problem by leveraging the motion priors of video diffusion models. We propose a controllable diffusion-based image animation method capable of animating arbitrary objects within an image while preserving their details. Our practical experience has shown that creating animations solely through prompt text is laborious and challenging, and results in limited control over finer details. To enhance user control over the animation process, we introduce the motion area guidance and motion strength guidance, allowing for precise and interactive control of the motion speed of multiple objects, significantly improving controllable and fine-grained image animation.


To accurately identify movable objects and their corresponding movable regions within an image, we introduce motion area masks. Inspired by ControlNet~\cite{zhang2023adding}, we append the mask along the channel dimension of the video latent representation and initialize the convolutional weights to zero, allowing them to adjust incrementally during the training process. This approach enables fine-grained and precise control over multiple movable areas in the input image, even when using multiple prompting texts. Training the model to follow the guidance of the motion area mask presents a significant challenge, as it is difficult to amass and annotate a substantial corpus of real videos in which only specific regions are in motion. To address this issue, we propose an unsupervised technique to generate synthetic videos with motion area masks derived from actual videos. The model is trained on both synthetic and real videos to ensure the generation of realistic videos directed by the motion area mask.

To effectively control the speed of moving objects in image animation, we introduce the metric of motion strength. 
Frame rate per second (FPS) represents the number of frames displayed in one second and previous methods~\cite{hu2022make,wang2023videocomposer,Chen2023VideoCrafter1OD} relied on FPS to control motion speed. 
However, it is important to note that different object types may exhibit varying motion speeds, and FPS primarily serves as a global scaling factor to indirectly adjust the motion speed of multiple objects. For instance, a video featuring a sculpture may have a high FPS but zero motion speed. To enable direct learning of motion speed by the video diffusion model, we propose a novel motion strength loss to supervise the model in learning inter-frame variances in the latent space.



In recent developments, image-to-video diffusion models~\cite{ho2022video,wang2023videocomposer,Chen2023VideoCrafter1OD} have emerged for generating realistic videos from input images. 
These models employ the CLIP vision encoder~\cite{radford2021learning} to encode the reference image, capturing its semantic structure and style.
However, they often struggle to preserve fine-grained visual details of the reference image. 
To address this limitation and apply video diffusion models to image animation tasks, we propose encoding the reference image into a latent space using a VAE~\cite{kingma2013auto}. 
This latent representation can then serve as the first frame of the generated video, effectively preserving image details without introducing additional parameters to existing video diffusion models, thereby enabling efficient training and inference for image animation tasks.


The combination of prompt text, motion area mask, and motion strength guidance enables the generation of complex image animations in diverse real-world scenarios. A sequential approach can be adopted, where a specific object is initially animated with a prompt text, followed by the animation of another object with a different prompt text. This iterative process allows users to progressively and interactively modify the image animation until a satisfactory outcome is achieved. An example of interactive image animation is illustrated in the last row of Figure~\ref{fig:it2v}. Initially, the actions of the boy are manipulated using a motion area mask marked in red together with the prompting text: ``the boy is running on the beach''. Subsequently, one of the two girls is animated through another motion area mask marked in green, accompanied by the prompting text: ``the girl runs to the sea''.
In summary, our contributions in this work are threefold:
\begin{itemize}
\item We introduce a highly flexible motion guidance mechanism that allows for fine-grained open domain image animation guided by motion area and motion strength. We propose a synthetic motion area mask generation method and a novel motion strength loss for effective training. 
\item Our image animation framework supports combined guidance from text, motion area, and motion strength, enabling an interactive animation generation process for creating complex animations.
\item By enhancing the image guidance mechanism of the current video diffusion model, we achieve a state-of-the-art FVD score of 443 on the zero-shot MSRVTT dataset.
\end{itemize}




%% file: sec/2_related.tex
\section{Related Work}
\begin{figure*}[t]
    \centering

    \includegraphics[width=\textwidth]{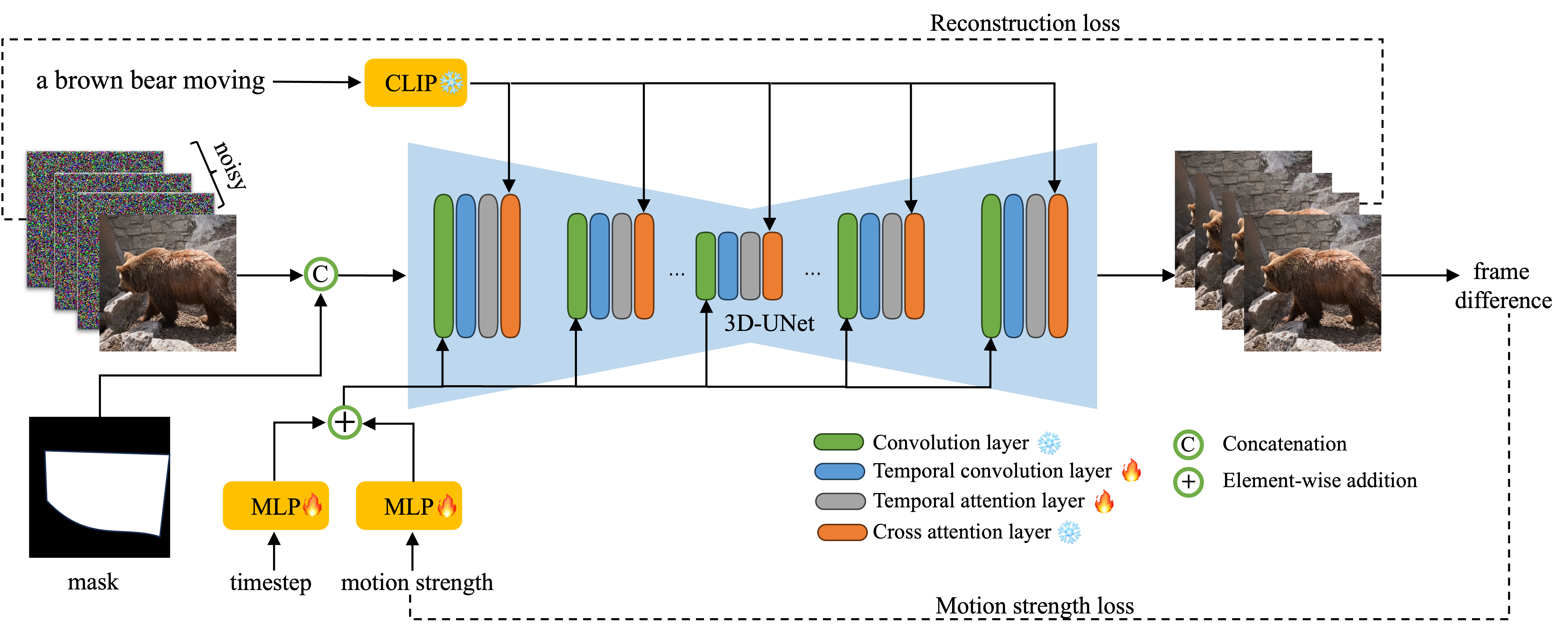}
    \vspace{-3mm}
    \caption{
    Overview of our pipeline. We adopt the widely used 3D U-Net based video diffusion model~\cite{esser2023structure,wang2023videocomposer} for image animation. Given a noisy video latent with shape (frames, height, width, channel), we concatenate the clean latent of the reference image and the noisy frames in the temporal dimension. Additionally, we concatenate the motion area mask with the video latent in the channel dimension. This results in the input latent with shape (frames+1, height, width, channel+1) for the 3D U-Net. To control the motion strength of the generated video, we project the motion strength as positional embedding and concatenate it with the time step embedding.}
    \label{fig:pipeline}
    \vspace{-5mm}
\end{figure*}

\subsection{Image Animation}
Image animation involves creating a video sequence from a static image. Initial approaches manually simulated motion for specific object types, such as mass simulation for vegetation and 2D rigid transformations for inanimate objects \cite{chuang2005animating, jhou2015animating}. Later research predicted inter-frame flows and utilized neural rendering to produce cinemagraphs, primarily focusing on fluid dynamics \cite{holynski2021animating,mahapatra2022controllable,okabe2009animating,Li20233DCF,Mahapatra2023SynthesizingAC}, natural environments \cite{xiong2018learning,li2023generative,cheng2020time,jhou2015animating,shaham2019singan}, and human features such as hair \cite{xiao2023automatic}, faces \cite{wang2021latent,wang2020imaginator,geng2018warp}, and body movements \cite{wang2021latent,weng2019photo,karras2023dreampose,siarohin2021motion,blattmann2021understanding,bertiche2023blowing}. Recent advancements have seen diffusion-based models that generate diverse video content guided by a still image.
The Make-it-move framework \cite{Hu2021MakeIM} captures motion patterns through motion anchors, which are then used to condition a VQ-VAE model. Despite its innovation, the model is limited by its dependency on the specific motion-object pairs from the training data, restricting its efficacy on open-domain images. The LFDM model \cite{ni2023conditional} adopts a two-stage process for human-centric video generation, while Generative Dynamics \cite{li2023generative} focuses on modeling oscillatory motion in natural scenes. However, these methods are domain-specific. In contrast, our approach addresses the challenge of open-domain image animation.






\subsection{Image generation with diffusion models}
The evolution of image generation research has transitioned from traditional frameworks such as Generative Adversarial Networks (GANs) \cite{goodfellow2020generative}, Variational Autoencoders (VAEs) \cite{kingma2013auto}, and autoregressive transformer models (ARMs) \cite{chen2020generative}, to the more recent diffusion models (DMs) \cite{ho2020denoising}. This shift is attributed to the stability, superior sample quality, and conditional generation capabilities of DMs. DALLE-2 \cite{ramesh2022hierarchical} represents a significant advancement by integrating the CLIP model \cite{radford2021learning} for text-image feature alignment, enabling text-prompted image synthesis. GLIDE \cite{nichol2021glide} introduces classifier-free guidance to refine image quality, while Imagen \cite{saharia2022photorealistic} utilizes a sequence of diffusion models for high-resolution image creation. Latent Diffusion Models (LDMs) \cite{ramesh2022hierarchical} employ an autoencoder \cite{esser2021taming} to manage the diffusion process in latent space, enhancing efficiency. Subsequent models like DiTs \cite{peebles2023scalable} and SDXL \cite{podell2023sdxl} further concentrate on latent space manipulation.
For conditional generation, T2I-Adapter \cite{mou2023t2i} and ControlNet \cite{zhang2023adding} have been developed to integrate spatial conditions such as depth maps and sketches into LDMs. Composer \cite{huang2023composer} extends this by training LDMs with multiple conditions for more precise control. Building on these developments, our work introduces motion area and motion strength guidance to provide fine-grained control of image animation.

\subsection{Video generation with diffusion models}

Recent advancements in diffusion models (DMs) have shown great promise in video generation \cite{nichol2021glide, saharia2022photorealistic, rombach2022high, peebles2023scalable, podell2023sdxl}. The Video Diffusion Model (VDM) \cite{ho2022video} pioneers this domain by adapting the image diffusion U-Net architecture \cite{ronneberger2015u} into a 3D U-Net structure for joint image and video training. Imagen Video \cite{ho2022imagen} employs a series of video diffusion models for high-resolution, temporally coherent video synthesis. Make-A-Video \cite{singer2022make} innovatively learns motion patterns from unlabeled video data, while Tune-A-Video \cite{wu2023tune} explores one-shot video generation by fine-tuning LDMs with a single text-video pair. Text2Video-Zero \cite{khachatryan2023text2video} tackles zero-shot video generation using pretrained LDMs without further training. ControlVideo \cite{zhang2023controlvideo} introduces a hierarchical sampler and memory-efficient framework to craft extended videos swiftly. Concurrently, VideoCrafter1 \cite{Chen2023VideoCrafter1OD} integrates image guidance from CLIP into diffusion models via cross-attention.
Despite these innovations, capturing complex motion and camera dynamics remains challenging. VideoComposer \cite{wang2023videocomposer} and DragNUWA \cite{yin2023dragnuwa} propose motion trajectory-based control for video generation, yet they fall short in interactive animation with multiple objects. Our approach addresses this by incorporating motion masks and motion strength parameters to precisely manipulate individual objects within an image, thus facilitating more user-centric interactive animation generation.


%% file: sec/3_method.tex
\section{Method}
We provide an overview of the video diffusion model in Section~\ref{sec:background}, followed by its adaptation for the image animation task in Section~\ref{sec:i2v}. Sections~\ref{sec:motion_mask} and~\ref{sec:motion_strength} offer detailed information on the integration of motion guidance into the generation process. Lastly, Sections~\ref{sec:guidance} and~\ref{sec:inference} present the inference process under different guidance.

\subsection{Background}\label{sec:background}

In this section, we introduce the preliminary knowledge of the latent diffusion-based model (LDM)~\cite{rombach2022high}. Given an image sample $x_0 \in \mathbb{R}^{3 \times H \times W}$, the LDM initially utilizes a pre-trained VAE to encode $x_0$ into a down-scaled latent representation $z_0 \in \mathbb{R}^{c \times h \times w}$. 
The forward process of the LDM can be described as a Markov chain that incrementally introduces Gaussian noise to the latent representation:
\begin{equation}
    q(z_t|z_{t-1}) = \mathcal{N}(z_t; \sqrt{1-\beta_t}z_{t-1}, \beta_tI),
\end{equation}
where $t=1,...,T$ and $T$ denotes the total number of timesteps. $\beta_t$ is a coefficient that controls the noise strength in step $t$.
The iterative noise adding can be simplified as:

\begin{equation}
    z_t=\sqrt{\bar{\alpha}_t}z_0 + \sqrt{1 - \bar{\alpha}_t}\epsilon, \quad\epsilon \sim \mathcal{N}(0, I),
    \label{eq:add_noise}
\end{equation}
where $\bar{\alpha}_t = \prod_{i=1}^t(1-\beta_t)$. During training, the LDM learns the latent space distribution of the real data by predicting the noise $\epsilon$ added on $z_t$, resulting in a reduction of computational complexity for diffusion models. The objective function can be written as:
\begin{equation}
    l_\epsilon = ||\epsilon - \epsilon_\theta(z_t, t, c)||^2_2,
    \label{eq:training_objective}
\end{equation}
where $\epsilon_\theta(\cdot)$ denotes the noise prediction function of diffusion models, which is implemented by an U-Net \cite{ronneberger2015u} architecture. To control the generation process flexibly, LDM employs a domain-specific encoder to map the user-input condition $c$ into an intermediate representation, which is then injected into the UNet via a cross-attention layer.



Video diffusion models~\cite{ho2022video, singer2022make, esser2023structure, wang2023videocomposer} expand upon the image LDM by incorporating a 3D U-Net, enabling them to effectively handle video data. The 3D U-Net incorporates an extra temporal convolution following each spatial convolution and a temporal attention block following each spatial attention block. To inherit the generation capacity from image data, the 3D U-Net is trained concurrently with both image and video data.
\begin{figure*}
    \centering
    \includegraphics[width=\textwidth]{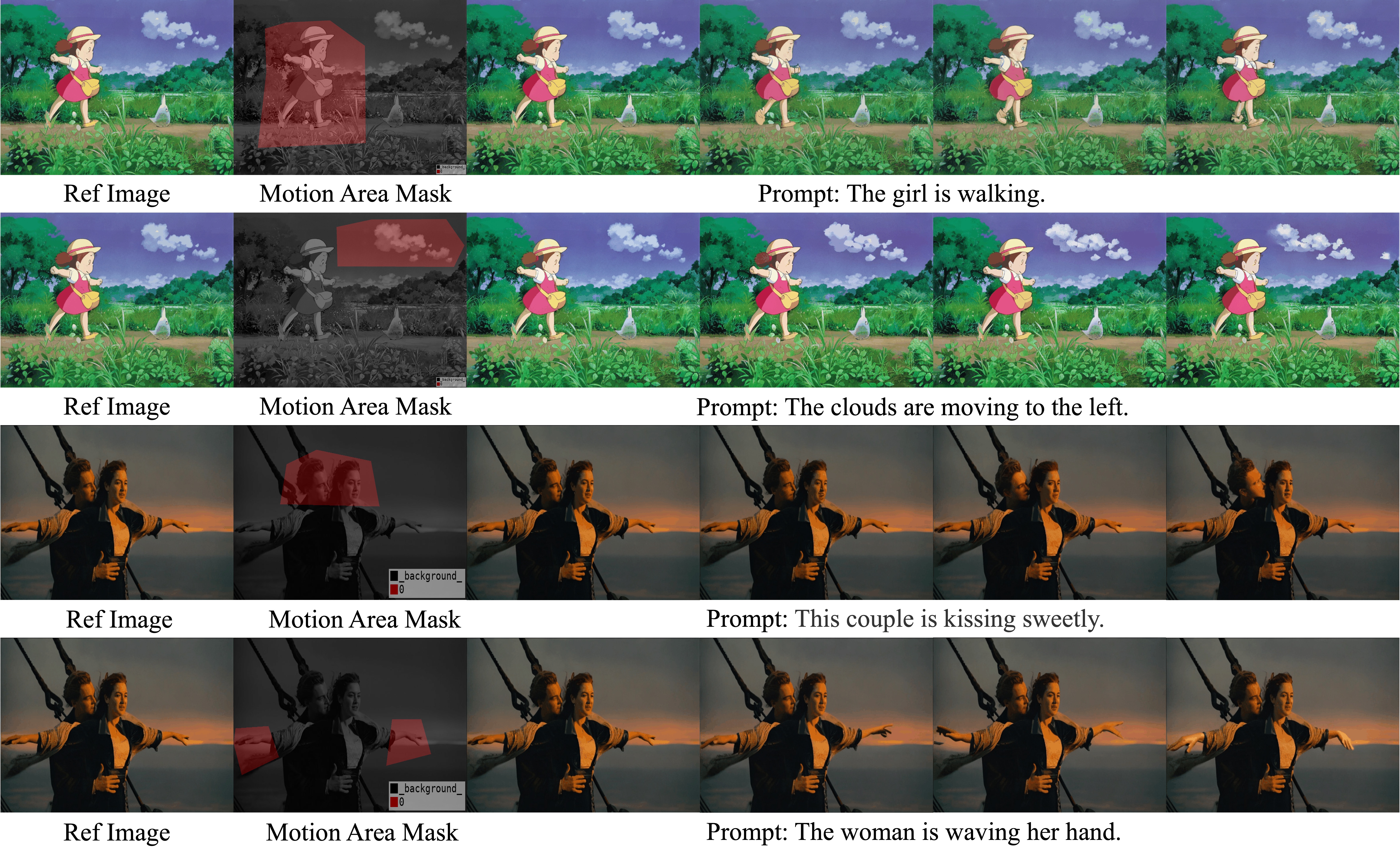}
    \vspace{-3mm}
    \caption{Motion mask guidance examples. The first column and second column are the input mask and motion mask respectively. The user can specify one or multiple movable areas in the motion mask to fine grained control the video generation.}
        \vspace{-5mm}
    \label{fig:mask_demo}
\end{figure*}
\subsection{Image Animation with Video Diffusion Model}\label{sec:i2v}

We employ the LDM VAE~\cite{rombach2022high} to encode the reference image into a latent representation, denoted as $z_{ref}$, in order to retain more appearance details. The VAE is trained for image reconstruction, thus $z_{ref}$ contains rich low-level image features. Although it may contain less semantic information compared to CLIP~\cite{radford2021learning} vision tokens, the diffusion model itself has demonstrated powerful semantic understanding capabilities in tasks such as semantic segmentation~\cite{Karazija2023DiffusionMF, Li2023OpenvocabularyOS}. As illustrated in Figure~\ref{fig:pipeline}, our training pipeline uses the reference image as the initial frame and adopts an auto-regressive strategy to forecast subsequent frames, facilitating image animation without extra model parameters. The first frame's content is propagated to later frames via temporal convolution and attention mechanisms. Consequently, only the temporal layers are fine-tuned, while the spatial layers remain frozen. 
At each time step 
$t$, we concatenate the clean $z_{ref}$ with the noisy latent $z_t$, which contains $N$ frames, resulting in a latent representation with $(N+1)$ frames as input. 
We then select only the last $N$ frames from the denoised $z_t$.


\subsection{Motion Area Guidance}\label{sec:motion_mask}
we introduce motion area guidance to provide users with precise control over the movable area of the input image. As shown in Figure~\ref{fig:pipeline}, we concatenate the motion area mask with the video latent in the channel dimension. Drawing inspiration from ControlNet~\cite{zhang2023adding}, we initialize the convolution kernel of the mask channel with zeros to maintain the original video generation capability.

We construct the training pairs of video and corresponding motion area masks from real videos using the following approach. Firstly, we convert the given video sample with $N$ frames to gray-scale. Then, we calculate frame differences that exceed a threshold value $T_m$. These differences are then combined to create the difference binary mask $d$:
\begin{equation}
 d = \bigcup\limits_{i=1}^{N-1} (|x_{gray}^i-x_{gray}^{i-1}| > T_m),
\end{equation}
where $x_{gray}^i$ is the gray-scale of the $i$-th frame, the threshold $T_m$ determines the intensity of motion in both movable and non-movable areas. If $T_m$ is set too high, objects in non-movable areas may still appear to move. Conversely, if $T_m$ is set too low, objects in non-movable areas might be completely frozen, potentially causing image artifacts near the boundary of the motion area mask. 
Subsequently, we identify the contours of these difference areas in $d$ and construct the motion area mask $m$ by assigning label 1 to the pixels contained within these contours, indicating the movable area. Finally, considering the motion area mask $m$, we post-process the video latent $z_0$ such that pixels in the non-movable area are reset to the values of the first frame $z_0^0$:
\begin{equation}
z_0' = (1-m) \cdot z_0^0 + m \cdot z_0.
\end{equation}
We use $z_t^i$ to denote the $i$-th frame of the video latent at time step $t$. 
As illustrated in Section~\ref{sec:ablation}, the post-processing step significantly enhances the effectiveness of the motion area guidance. 
To address subtle movements imperceptible to the human eye, which should not be marked as the movable area, we explicitly instruct the model to keep these pixels unchanged. 
The motion threshold $T_m$ is adjusted to ensure that the visual differences between the reconstructed video $z_0'$ and $z_0$ remain reasonably small. 
The impact of motion area guidance on is demonstrated in Figure~\ref{fig:mask_demo}.

\subsection{Motion Strength Guidance}\label{sec:motion_strength}
\begin{figure}
    \centering
    \includegraphics[width=0.48\textwidth]{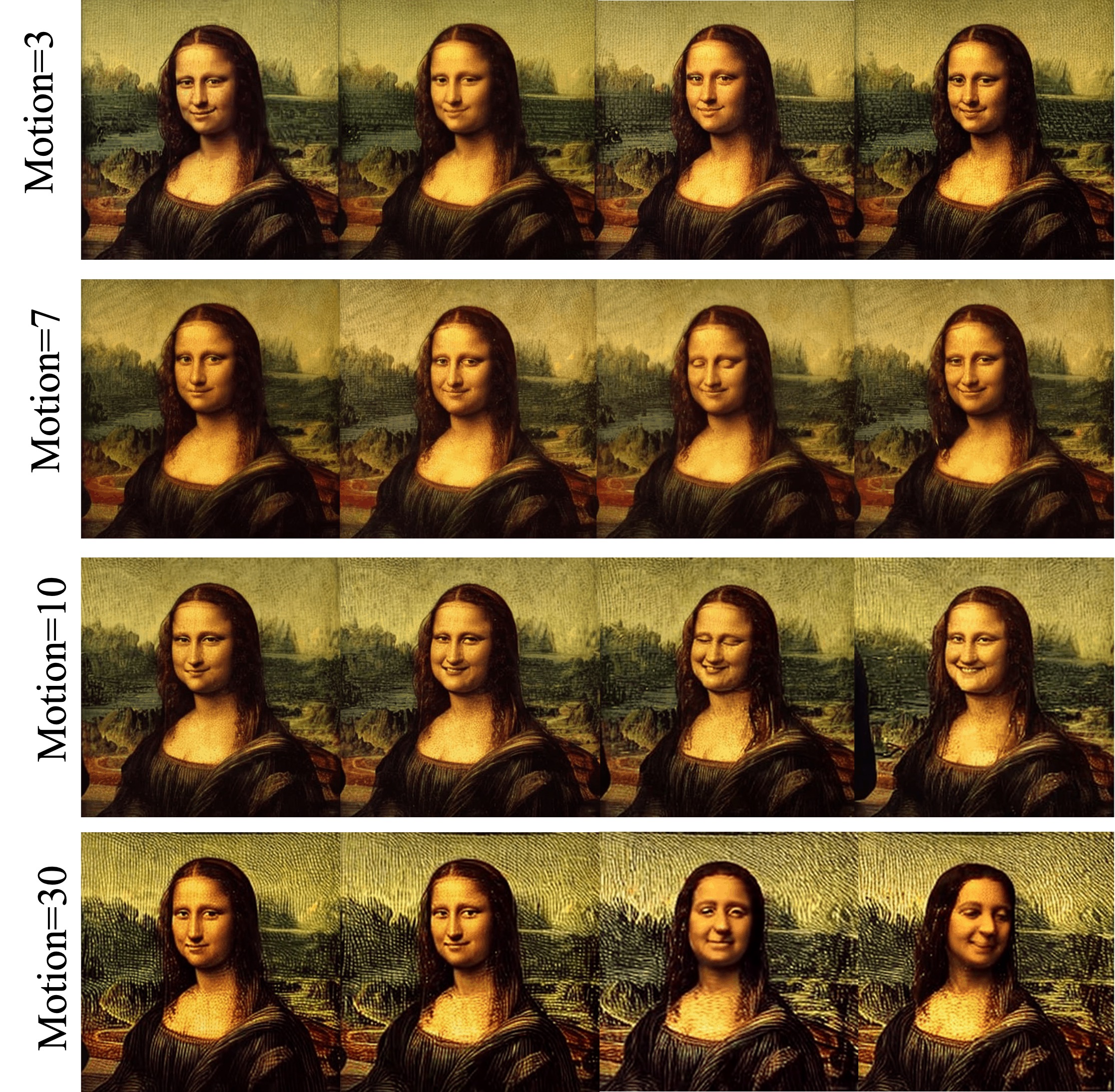}
    \vspace{-3mm}
    \caption{Motion strength guidance examples. Augmenting the motion strength accelerates the alteration of Mona Lisa's expression, but excessive motion strength may lead to the loss of fine-grained facial details.}
    \vspace{-3mm}
    \label{fig:motion_compare}
\end{figure}

During our training process, we observed that the sampling FPS affects the motion speed of the movable objects in the generated videos. However, using FPS alone as the motion speed guidance for video generation is inadequate, as videos with the same FPS may exhibit varying motion speeds based on their content. Therefore, FPS alone cannot effectively regulate the speed of object motion. Consequently, we propose a metric called motion strength $s$ to quantitatively measure the motion speed of the target motion area:
\begin{equation}
s(z) = \frac{1}{N-1}\sum_{i=1}^N |z^i - z^{i-1}|.
\end{equation}
Here, the motion strength quantifies the differences between frames in the latent space. 
Similar to the time step, we project the motion strength into a positional embedding and add it to each frame in the residual block to ensure uniform application of the motion strength to every frame. 
The impact of motion strength guidance on the animation results is illustrated in Figure~\ref{fig:motion_compare}.

Training our motion strength guided pipeline directly with the noise prediction loss defined in Equation~\ref{eq:training_objective} is hard to converge. 
This is likely due to the fact that the noise prediction loss is primarily influenced by the frame-level image difference and does not directly supervise the inter-frame difference. 
Therefore, we introduce a motion strength loss to directly supervise the inter-frame difference:
\begin{equation}
l_{s} = ||s(z_0) - s(\hat{z}_0)||^2_2 , 
\end{equation}
where $\hat{z}_0$ represents the model's estimated clean video latent $z_0$, which can be obtained by transforming Equation~\ref{eq:add_noise} as:
\begin{equation}
\hat{z}_0 = \frac{z_0 - \sqrt{1 - \bar{\alpha}_{t}} ~ \epsilon_\theta(z_t, t, c)}{\sqrt{\bar{\alpha}_t}}.
\end{equation}
Finally, we combine the noise prediction loss and the motion strength loss with the scaling factor $\lambda$.
\begin{equation}
l = l_{\epsilon} + \lambda \cdot l_{s}.
\end{equation}

\subsection{Guidance Composition}\label{sec:guidance}
Our image animation model integrates guidance from reference images, text, motion areas, and motion strength. During training, we vary the textual prompt and motion area to allow the model to accept different combinations of guidance during inference. However, conflicting guidance inputs can diminish their individual effects. For instance, if the text prompt does not align with the content of the reference image, the model prioritizes fidelity to the image. Fortunately, with motion area guidance, objects outside the motion area mask are completely frozen, allowing for interactive editing of the generated animation. As demonstrated in the last row of Figure~\ref{fig:it2v}, different objects can be animated with different texts.



\subsection{Shared Noise Inference}\label{sec:inference}
During training, we construct the input latent by adding noise on the clean video latent. This noise schedule leaves some residual signal even at the terminal diffusion timestep T. As a consequence, the diffusion model fails to generalize faithful image animation during test time when we sample from random Gaussian noise without any real data signal. To solve this train-test discrepancy problem, during testing, we obtain the base noise by adding noise on $z_{ref}$ using the forward process of DDPM~\cite{ho2020denoising}. 
As the DDPM independently samples random noise $\epsilon^i$ for each frame, it allows for frame diversity. 
The DDPM independently samples random noise $\epsilon^i$ for each frame, allowing for frame diversity.
The noise latent for frame $i$ can be expressed as:
\begin{equation}
z_T^i = \sqrt{\bar{\alpha_T}} z_{ref} + \sqrt{1-\bar{\alpha_T}}\epsilon^i,
\end{equation}
where $\bar{\alpha_T}$ denotes the diffusion factor. 
This approach combines the base noise with $z_{ref}$ to achieve a balance between preserving the reference image information and introducing frame-specific diversity through the random noise $\epsilon^i$. This design decision is critical for high image fidelity animation.

%% file: sec/4_exp.tex
\section{Experiment}\label{sec:exp}
\begin{figure*}
    \centering
    \includegraphics[width=0.98\textwidth]{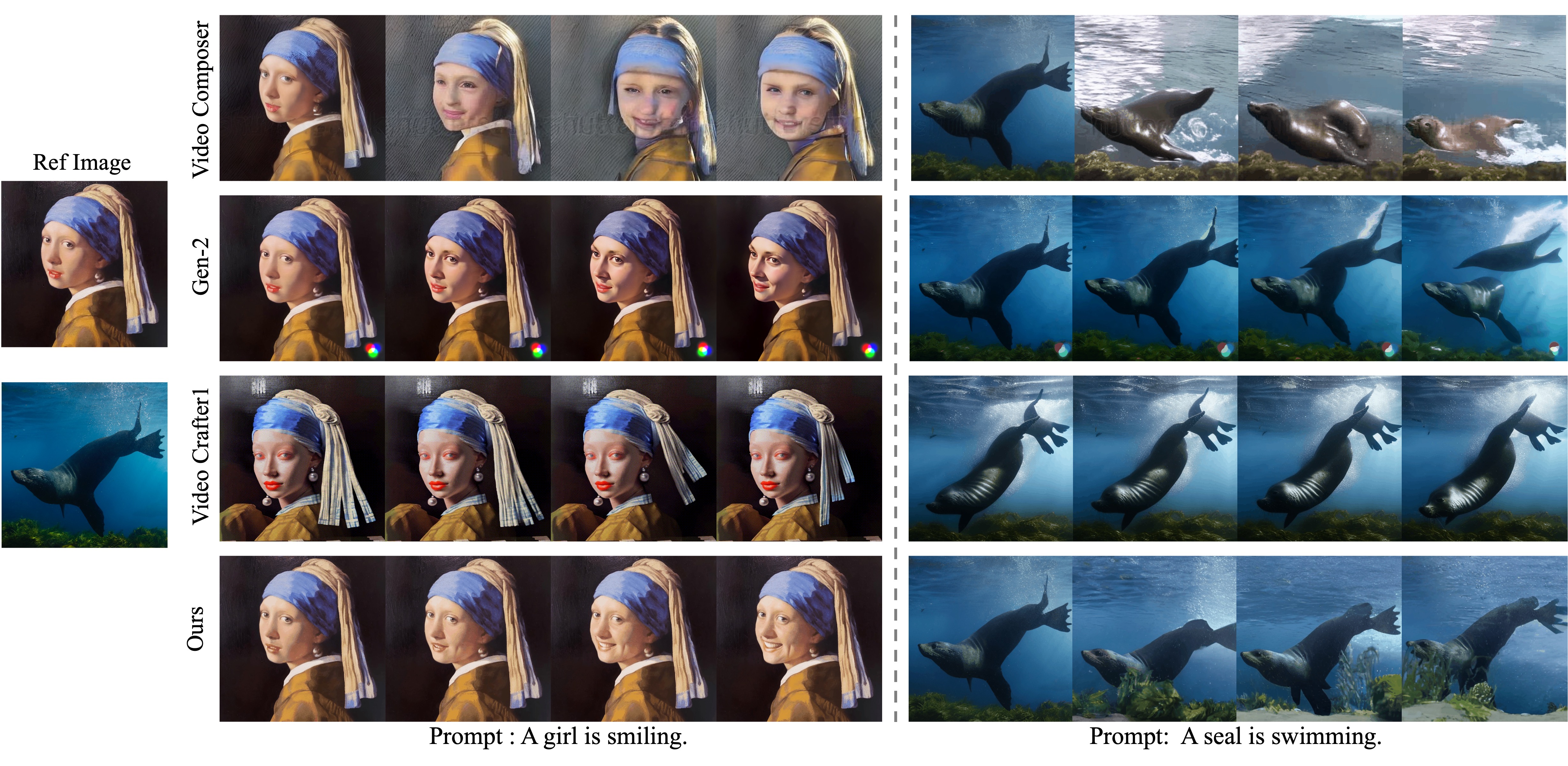}
    \vspace{-2.5mm}
    \caption{Qualitative Results. Comparing to open sourced methods Video Composer and VideoCrafter1, our method achieves higher image fidelity. Comparing to the commercial product Gen-2, our method achieves higher frame consistency.}
    \vspace{-5mm}
    \label{fig:i2v_compare}
\end{figure*}
\subsection{Experimental Setup}
\noindent \textbf{Datasets.} 
Our model is initialized from VideoComposer~\cite{huang2023composer} which pretrains on WebVid10M~\cite{bain2021frozen}. Then we finetune it on 20K videos randomly sampled from HD-VILA-100M~\cite{xue2022hdvila} to remove watermark. 
Following ~\cite{wu2022nuwa,singer2022make,wang2023modelscope}, we conduct evaluation on MSR-VTT dataset~\cite{xu2016msr} in a zero-shot setting. MSR-VTT is an open domain video retrieval dataset, where each video clip is accompanied by 20 natural sentences for description. Typically, the textual descriptions corresponding to the 2,990 video clips in its test set are utilized as prompts to generate videos. 

\noindent \textbf{Evaluation Metrics.}
Following previous methods~\cite{singer2022make, esser2023structure}, we use Frechet Video Distance (FVD) \cite{unterthiner2019fvd} to measure the video generation quality. We also use the Frame Consistency measured via CLIP cosine similarity of consecutive frames to show the temporal consistency of video. We propose the metric Motion Mask Precision to measure the effect of motion mask guidance, which calculates the percentage of the moving area of generated video within the given motion area mask. To measure the effect of motion strength guidance, we propose Motion Strength Error, which is the mean squared error of the generated video motion strength and the given video motion strength. 
In this section, all statistics are collected by generating 16-frame videos in 256 $\times$ 256 resolution with DDIM~\cite{song2021denoising} algorithm in 50 steps.

\noindent \textbf{Implements Details.}
We employ the AdamW \cite{DBLP:conf/iclr/LoshchilovH19} optimizer with a learning rate of $5 \times 10^{-5}$ for training our model. All experiments are conducted on a single NVIDIA A10 GPU, requiring approximately 20 GB of vRAM for training and 6 GB of vRAM for inference. To enhance the model performance, we conduct multiple frame rate sampling during training, utilizing various frame rates (e.g., 4, 8, 12) to obtain an 8-frame training clip  with a resolution of 384 $\times$ 384. We train the model for 10,000 iterations with a batch size of 2, which costs about 1 day. 
The motion area mask threshold $T_m$ is set to 5 and the motion strength loss scaling factor $\lambda$ is set to 0.001.

\subsection{Quantitative Results}
Table \ref{tab1} presents the quantitative results of the zero-shot video generation ability on MSR-VTT compared to existing methods. To ensure a fair comparison, all models are evaluated on a resolution of 256 $\times$ 256. With the same text and first frame condition, our method achieves lower FVD comparing to VideoComposer and concurrent work VideoCrafter1, which is evident that integrating fine-grained spatial information into the conditions leads to a significant improvement. 
This improvement suggests that our method is capable of generating more coherent videos than previous approaches.

\begin{table}[h]\small
\begin{tabular}{cccc}
\hline
Method         & Conditions & Params(B) & FVD($\downarrow$)  \\ \hline
CogVideo~\cite{hong2022cogvideo}         & text    & 15.5      & 1294                     \\
LVDM~\cite{he2022latent}               & text    & \textbf{1.16}      & 742                      \\
MagicVideo~\cite{zhou2022magicvideo}        & text    & -         & 998                     \\
VideoComposer~\cite{wang2023videocomposer}    & text    & 1.85      & 580                      \\
VideoFusion~\cite{luo2023videofusion}       & text    & 1.83      & 581                     \\
ModelScope~\cite{wang2023modelscope}       & text    & 1.70      & 550                      \\
VideoComposer~\cite{wang2023videocomposer}    & text\&image    & 2.49      & 551                      \\
VideoCrafter1~\cite{Chen2023VideoCrafter1OD}  & text\&image    & 3.24      & 465   \\
Ours              & text\&image    &   1.81      & \textbf{443}        \\ \hline
\end{tabular}
\caption{Video generation performance on the test set of MSR-VTT. “Conditions” denotes the type of condition for generation.}
\vspace{-1mm}
\label{tab1}
\vspace{-1mm}
\end{table}


\subsection{Qualitative Results}

We present several visual examples of our method with three baselines: VideoComposer~\cite{wang2023videocomposer}, Gen-2~\cite{Gen2} and VideoCrafter1~\cite{Chen2023VideoCrafter1OD} in Figure~\ref{fig:i2v_compare}. All the methods have the ability to generate videos based on reference image, thus showcasing the significant progress in image-conditioned video generation. VideoComposer and VideoCrafter1 achieves satisfactory levels of fluency but lost details of the reference image.
In contrast, our proposed method preserves the latent representation of the first frame and denoises the subsequent frames, which plays a fundamental role in ensuring faithful video generation. Despite Gen-2's relatively successful preservation of temporal consistency, it falls short in effectively controlling the motion amplitudes of the generated videos. In contrast, our method offers the flexibility of adjusting motion strength, where larger strengths correspond to greater motion amplitudes.

%% file: sec/5_ablation.tex
\subsection{Ablation Study}\label{sec:ablation}

\paragraph{Image condition}
We conduct an analysis of various design strategies for injecting reference image information in Table~\ref{tab:image_condition}. The ``CLIP Vision Global Token" method used by Video Composer~\cite{wang2023videocomposer} encodes the input image to vision tokens using CLIP image encoder and only injects the global [CLS] token into the U-Net through cross attention. The ``CLIP Vision Full Tokens" method, used by VideoCrafter1 \cite{Chen2023VideoCrafter1OD}, projects and concatenates all vision tokens with the text embedding. As shown in Figure \ref{fig:i2v_compare}, CLIP vision tokens contain rich semantic information by pretraining on text and image pairs, but image details such as background texture and face details may be lost. Using all vision tokens also requires more computation cost to compute the cross attention between the latent and additional vision tokens. To preserve more image details, we first try the ``Concat Latent Spatial" method, which concatenates the reference image latent from VAE with every frame on the channel dimension. However, we notice that this method limits the diversity of video motion when generating long videos. Finally, we concatenate the input image latent and the noise latent on the temporal dimension. As shown in ``Concat Latent Temporal", this method produces videos with better image fidelity and achieves better inference efficiency without using the additional ViT vision encoder.
\begin{table}\small
    \centering
    \setlength{\tabcolsep}{0.5mm}{
    \begin{tabular}{l|c|c|c|c}
    \hline
    Method & FVD & Consistency & Time(s) & Mem(G) \\
    \hline
    CLIP Vision Global Token & 551 & 0.879 & 22.13 & 11.1 \\
    CLIP Vision Full Tokens &457 & 0.911 & 24.12  & 11.7\\
    Concat Latent Spatial & 450 & 0.917 &  20.38 & 5.7 \\
    Concat Latent Temporal & 443 & 0.916 & 21.54 & 5.9 \\
    \hline
    \end{tabular}
    }
    \caption{The performance comparison of image condition designs.}
    \vspace{-3mm}
    \label{tab:image_condition}
\end{table}

\paragraph{Motion area guidance}\vspace{-3mm}
Table \ref{tab:mask_condition} illustrates different design choices for motion area guidance. A simple and training-free approach to freeze the area outside the motion area is to apply the motion mask on the noise video latent, therefore the latent values outside motion area are to the same. However, this method produces videos nearly not moves, since the freeze noise latent is not Gaussian noise, which is inconsistent with the training setting.  In Table \ref{tab:mask_condition}, 
the "Mask Guidance" method concatenates the motion area mask and the noise latent as the input to the U-Net. "Mask Guidance + Freeze" is our proposed method that the latent values in non-movable areas are frozen and set to be the same as the first frame. We can observe the strategy of freezing non-movable areas helps the model to easily learn the guidance between the motion area mask and the target video.
\begin{table}\small
    \centering
    \begin{tabular}{l|c}
    \hline
    Method & Motion Mask Precision \\
    \hline
    No Control & 0.21 \\
    Mask Guidance & 0.52\\
    Mask Guidance + Freeze & 0.82\\
    \hline
    \end{tabular}
    \caption{Ablation study of the motion mask guidance. Our proposed strategy demonstrates effective capability in following designated motion areas to generate corresponding animation videos.}
    \label{tab:mask_condition}
\end{table}

\begin{table}\small
    \centering
    \begin{tabular}{l|c}
    \hline
    Method & Motion Strength Error \\
    \hline
    No Control & 14.19  \\
    Adding Noise & 12.74  \\
    FPS Guidance & 8.37 \\
    Motion Strength Guidance &  4.82  \\
    Motion Strength Guidance + Loss & 2.36\\
    \hline
    \end{tabular}
    \caption{Ablate the design of the motion strength guidance. Comparing to FPS guidance, our method offers greater flexibility in incorporating motion into animation videos.}
    \label{tab:strength_condition}
    \vspace{-3mm}
\end{table}
\paragraph{Motion strength guidance}\vspace{-3mm}
We compare different design choices of motion strength guidance in Table~\ref{tab:strength_condition}. Since the initial noise video latent is generated by adding noise on $z_{ref}$, we can perform limited control over motion strength by the amount of added noise, as indicated by ``Adding Noise''. The performance of FPS guidance is largely depend on the video content, so its variance is large. Comparing to FPS, our motion strength guidance performs more stable and achiever lower motion strength error. 

%% file: sec/6_conclusion.tex
\section{Conclusion and Future Work}
In this paper, we have presented an open domain image animation pipeline with a video diffusion model, incorporating motion area and motion strength guidance. 
While our method has shown promising results in achieving fine-grained and interactive image animation, it is important to acknowledge that our model has not been trained on high-resolution videos due to limited training resources. 
This limitation constrains the applicability and performance of our method in generating high-resolution animations. 
